\documentclass[letterpaper, 10pt, conference]{ieeeconf}
\IEEEoverridecommandlockouts
\usepackage[utf8]{inputenc}

\usepackage{comment}
\usepackage{xspace}
\usepackage{graphicx}
\usepackage{overpic}
\usepackage{svg}
\usepackage{xcolor}
\usepackage{bm}
\definecolor{gg}{RGB}{215, 25, 28}
\definecolor{Gray}{gray}{0.95}

\usepackage{tabularx}
\usepackage{multirow}
\usepackage{booktabs}
\usepackage{colortbl}

\newcolumntype{I}{>{\centering}p{0.138\textwidth}}
\newcolumntype{Y}{>{\raggedleft\arraybackslash}X}

\usepackage{etoolbox}
\makeatletter
\patchcmd{\@makecaption}
{\scshape}
{}
{}
{}
\newcommand{\onetagright}{\tagsleft@false}
\makeatother

\usepackage{amsmath}
\usepackage{amssymb}
\usepackage{amsthm}
\usepackage{siunitx}

\newtheoremstyle{main}
{1em}                                                %
{1em}                                                %
{\itshape}                                        %
{0pt}                                                %
{\scshape}                                           %
{\\*}                                                %
{2pt}                                                %
{\thmname{#1}\thmnumber{ #2}: \thmnote{\itshape #3}} %

\usepackage{environ}
\newcounter{problem}
\NewEnviron{problem}[1]{%
	\refstepcounter{problem}
	\begin{subequations}\label{#1}\begin{align}
			\BODY
		\end{align}\end{subequations}
}

\usepackage[linesnumbered,ruled,noend]{algorithm2e}
\newcommand{\removelatexerror}{\let\@latex@error\@gobble}

\usepackage[
	activate   = {true},
	protrusion = false,
	expansion  = true,
	kerning    = true,
	spacing    = true,
	tracking   = false,
	auto       = true,
	selected   = true,
	factor     = 1000,
	stretch    = 20,
	shrink     = 20,
]{microtype}

\usepackage{csquotes}
\usepackage[
	maxbibnames=99,
	maxcitenames=2,
	natbib=true,
	style=numeric-comp,
	backend=biber,
	sorting=none,
	giveninits=true,
	url=false,
	doi=false,
	eprint=false,
	isbn=false,
]{biblatex}

\makeatletter
\let\NAT@parse\undefined
\makeatother
\usepackage[pdfa,colorlinks,bookmarksopen,bookmarksnumbered,allcolors=gg]{hyperref}
\usepackage[english]{babel}

\usepackage[nameinlink,capitalise]{cleveref}
\crefname{line}{line}{lines}
\crefname{figure}{Fig.}{Figs.}
\Crefname{figure}{Fig.}{Figs.}
\crefname{equation}{Eq.}{Eqs.}
\Crefname{equation}{Eq.}{Eqs.}
\crefname{section}{Sec.}{Secs.}
\Crefname{section}{Sec.}{Secs.}
\crefname{definition}{Def.}{Defs.}
\Crefname{definition}{Def.}{Defs.}
\crefname{algorithm}{Alg.}{Algs.}
\Crefname{algorithm}{Alg.}{Algs.}
\crefname{assumption}{Asm.}{Asms.}
\Crefname{assumption}{Asm.}{Asms.}
\crefname{subassumption}{Asm.}{Asms.}
\Crefname{subassumption}{Asm.}{Asms.}
\Crefname{problem}{Problem}{Problems}
\crefname{problem}{Problem}{Problems}

\newcommand{\mf}[1]{\mbox{\cref{#1}}\xspace}
\newcommand{\mfa}[1]{\mbox{\cref{#1}}}

\usepackage{flushend}

\let\labelindent\relax
\usepackage[inline]{enumitem}
\usepackage{mathtools}

\DeclarePairedDelimiterX{\infdivx}[2]{(}{)}{%
  #1\;\delimsize\|\;#2%
}

\newcommand{\cspace}{\mbox{\ensuremath{\mathcal{Q}}}\xspace}
\newcommand{\cspacespace}{\mbox{\ensuremath{\mathcal{Q}}}-space\xspace}
\newcommand{\cgoal}{\mbox{\ensuremath{\mathcal{Q}_{goal}}}\xspace}
\newcommand{\cfree}{\mbox{\ensuremath{\mathcal{Q}_{free}}}\xspace}
\newcommand{\prob}[1]{\ensuremath{\mbox{Pr}\left(#1\right)}}

\newcommand{\dof}{\textsc{d\scalebox{.8}{o}f}\xspace}
\newcommand{\ik}{\textsc{ik}\xspace}
\newcommand{\ccikop}{\textsc{ccikopt}\xspace}
\newcommand{\mnet}{\textsc{MotionBenchMaker}\xspace}

\newcommand{\eg}{\emph{e.g.},\xspace}
\newcommand{\ie}{\emph{i.e.},\xspace}

\graphicspath{ {./images/} }

\title{\LARGE \bf
Stochastic Implicit Neural Signed Distance Functions for Safe Motion Planning under Sensing Uncertainty
}
\author{Carlos Quintero-Peña, Wil Thomason, Zachary Kingston, Anastasios Kyrillidis and Lydia E. Kavraki%
\thanks{All authors are affiliated with the Department of Computer Science, Rice University, Houston TX, USA 
{\tt\small \{carlosq, wbthomason, zak, anastasios, kavraki\}@rice.edu}. This work was supported in part by NSF RI 2008720, NSF ITR 2127309 for the Computing Research Association CIFellows Project, and Rice University Funds.
}%
}

\begin{document}
\maketitle
\thispagestyle{empty}
\pagestyle{empty}

\begin{abstract}
Motion planning under sensing uncertainty is critical for robots in unstructured environments to guarantee safety for both the robot and any nearby humans.
Most work on planning under uncertainty does not scale to high-dimensional robots such as manipulators, assumes simplified geometry of the robot or environment, or requires per-object knowledge of noise.
Instead, we propose a method that \emph{directly} models sensor-specific aleatoric uncertainty to find safe motions for high-dimensional systems in complex environments, without exact knowledge of environment geometry.
We combine a novel implicit neural model of stochastic signed distance functions with a hierarchical optimization-based motion planner to plan low-risk motions without sacrificing path quality. %
Our method also explicitly bounds the risk of the path, offering trustworthiness.
We empirically validate that our method produces safe motions and accurate risk bounds and is safer than baseline approaches. %

\end{abstract}

\section{Introduction}\label{sec:introduction}

Robots in unstructured environments must reliably plan safe (\ie collision-free) motions using only uncertain, noisy sensor percepts.
For robots in human-oriented environments (\eg home or assistive robotics), this capability is crucial---as unsafe motions may hurt humans---and challenging, as these robots are often high degree-of-freedom (\dof) manipulators.
Reliable safety under uncertainty requires not only producing plans that are unlikely to collide, but also providing evidence that plans are trustworthy.
Moreover, for practical use, planners need to efficiently support complex environments without knowledge of the true environment geometry.

However, most work on motion planning under uncertainty makes simplifying assumptions about robot or environment geometry (\eg{} point robots or environments with only known, simple geometry)~\cite{dawson2020,blackmore2011_chance-constrained,lew20_chanceconstrained,luders10_chanceconstrained,summers18}, does not scale to high \dof systems, or places strict assumptions on the distributions of noise (\eg{} only translational noise, segmented to individual objects or normally distributed)~\cite{quinteropena2021_robustoptimization,dawson2020,dai19}.

In contrast, we introduce a method for reliable, safe motion planning for high \dof systems under sensing uncertainty that directly models inherent sensor noise without placing assumptions on the environment.
We propose to quantify the aleatoric uncertainty of the sensor with an implicit model of the stochastic signed distance fields between the robot's links and points in the environment, conditioned on the robot's configuration.
By explicitly modeling this uncertainty, we can both compute safe paths given only noisy sensing and approximately bound the remaining risk of collision.
\begin{figure}
    \centering
    \includegraphics[width=0.9\linewidth]{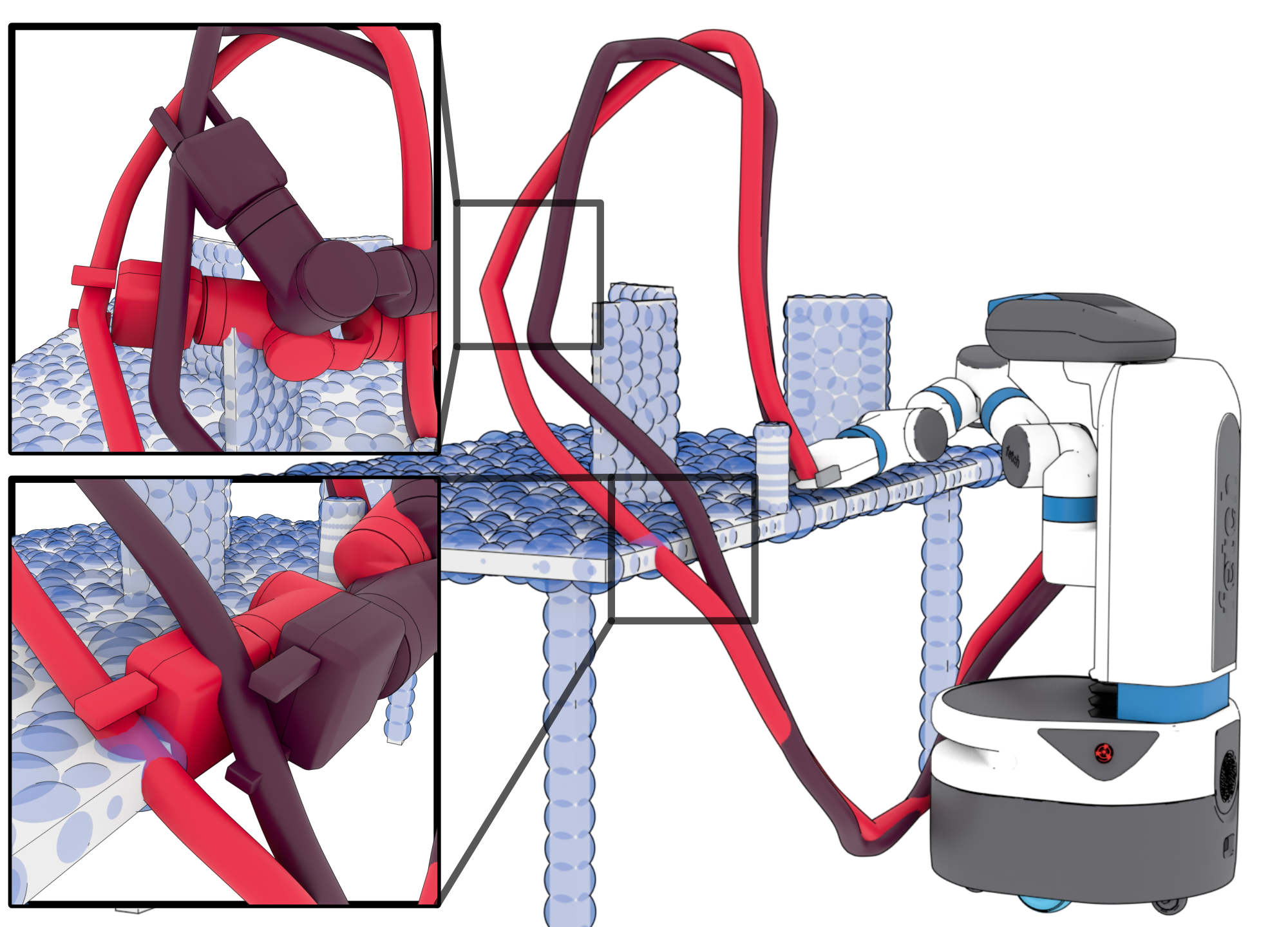}
    \caption{Simulated motion planning problem under sensing uncertainty. 
    The environment is composed of noisy points (blue spheres) to be avoided.
    The robot must plan to grasp the cylinder without colliding with the table or objects.
    Our method transforms a candidate path (red) into a safe path (purple) by solving a sequence of optimization problems that account for sensing uncertainty.
    Cutouts show parts of the path transformation: the arm is pushed away from noisy regions to attain safer behavior.}
    \label{fig:planning_framework2}%
\end{figure}

Specifically, we contribute
\begin{enumerate*}[label=(\arabic*)]
\item a variational inference perspective on modeling stochastic signed distance fields for motion planning (inspired by~\cite{shen2021_stochastic_neural}), used to learn
	\item an implicit neural model of sensor-specific noisy egocentric distance, which we incorporate in
	\item a novel chance-constrained inverse kinematics (\ik) formulation, allowing us to create
	\item a hierarchical planner that produces minimal risk motions (with respect to the learned distance model and an uncertainty-agnostic initial motion plan) in realistic environments
\end{enumerate*}.
Our learned model directly predicts distribution parameters for noisy distance measurements to arbitrary points in the environment, allowing it to capture the aleatoric uncertainty of the sensor in question without assuming that noise is segmented to the level of individual objects or requiring knowledge of object geometry.
We empirically validate that our model correctly predicts both distance values and their uncertainty, and that our planner finds motion plans that are both safe (\ie minimize risk) and reliable (\ie the predicted risk matches or conservatively upper-bounds the empirically measured probability of collision).
We further compare our planner to a commonly used baseline and show that, despite longer planning times, we produce significantly safer and higher-quality plans.\looseness=-1
\section{Preliminaries}

\label{sec:background_and_problem_definition}
We consider a robot with $n$ controllable joints and configuration space $\cspace \subseteq \mathbf{R}^n$.
We assume that the environment is represented by noisily measured 3D points corresponding to the external surfaces of objects. 
Point clouds~\cite{kuntz2020_fast} are an example of such a representation.
These coordinates are usually computed from depth information from, \eg{} a RGB-D camera or LiDAR, which is subject to imperfect measurements and other sources of errors.
The distance from the sensor to an object's surface can be modeled as a random variable with Gaussian distribution~\cite{khoshelham2012_accuracy}.
This source of sensing uncertainty tends to dominate in the settings we consider; the robot's proprioception (\ie{} via joint encoders) is typically much less noisy.\looseness=-1

In ``normal'' motion planning, we seek a collision-free path $\rho$ connecting the initial robot configuration $q_{start} \in \cspace$ to a goal region $\cgoal \subset \cspace$, \ie{} $\rho : [0,1] \rightarrow \cfree$, $\rho(0) = q_{start}, \rho(1) \in \cgoal$.
The goal of motion planning under sensing uncertainty is to find a path that is safe despite imperfect sensing information.
More specifically, we want a path whose probability of collision (\ie \textit{risk of collision}) is no larger than a given threshold $\Delta$.
This problem can be formulated as the chance-constrained optimization problem~\ref{prob:prob1}:
\begin{subequations}
\label{prob:chance_constrained_traj_formulation}
\begin{align}
\tag{Prob. 1}\label{prob:prob1}
& \underset{q_0, \dots, q_T}{\text{min}}
& & f(q_{0:T}) \nonumber \\
& \text{s.t.} 
& & q_0 = q_{\text{start}},\; q_T \in \cgoal, \nonumber\\
& & & q^{l} \leq q_t \leq q^{u}, \; t \in [0,\dots,T],  \nonumber\\
& & & \prob{\bigwedge_t q_t \in \cfree} \geq 1 - \Delta, \; \label{eq:chance_constrained_traj_nocollision}
\end{align}
\end{subequations}
where $q_0,\dots,q_T$ are waypoints of a discretized path, $f$ is the objective function (\eg to encourage smooth, short paths), and $q^l$ and $q^u$ are lower and upper joint limits.
\eqref{eq:chance_constrained_traj_nocollision} is a chance constraint enforcing that the probability of having no collisions \emph{along the path} remains above the threshold.
Unfortunately, this probability cannot be expressed in a tractable form suitable for optimization.

\section{Related Work}
\label{sec:related_work}

\subsection{Motion Planning under environmental uncertainty}\label{sec:motion_planning_under_environmental_uncertainty}

Collision chance constraints, or constraints on the probability that a robot's trajectory collides with a noisy environment, have been successfully used for safe motion planning under uncertainty by a wide range of work.
Chance constraints are typically determinized to keep the planning problem tractable.
These deterministic reformulations are then used by either optimization~\cite{blackmore06_aprobabilistic,blackmore2011_chance-constrained,lew20_chanceconstrained,dawson2020} or sampling~\cite{luders10_chanceconstrained,luders10_chanceconstrained,summers18,kajsa2023_distributionally}-based motion planners to generate provably safe trajectories.
Similarly, we also reformulate and enforce chance constraints to guarantee a desired maximum risk of collision.
For example,~\citet{blackmore06_aprobabilistic,blackmore2011_chance-constrained} create a disjunctive convex optimization problem that can be solved with branch-and-bound;  \citet{luders10_chanceconstrained} build a tree-like planner that validates states against the reformulated constraints.
\citet{summers18} uses a similar idea for non-Gaussian uncertainty and moment-based ambiguity sets of distributions.
Finally,~\citet{dawson2020} propose a differentiable surrogate risk for manipulator robots and convex obstacles under Gaussian translational uncertainty that is guaranteed to never underestimate the true risk, enforced by constraints in a nonlinear program.
Many of these methods rely on simplified robot shapes, \eg{} point robots~\cite{blackmore2011_chance-constrained,luders10_chanceconstrained,summers18,lew20_chanceconstrained}, obstacle shapes, \eg{} polyhedral~\cite{blackmore2011_chance-constrained,luders10_chanceconstrained,summers18} or convex~\cite{lew20_chanceconstrained,dawson2020}, and the noise model, \eg{} additive Gaussian noise on obstacle positions~\cite{blackmore2011_chance-constrained,luders10_chanceconstrained,dawson2020}.
In contrast, our method is designed for high-\dof robots and complex, noisy scenes, where point-robot assumptions are insufficient and strict assumptions on the noise distribution may not hold.

When reformulating chance constraints, most methods allocate equal risk for every waypoint and/or obstacle in the path to make the problem tractable~\cite{blackmore2011_chance-constrained,luders10_chanceconstrained,summers18,lew20_chanceconstrained}.
However, this strategy can lead to overly conservative solutions, since robot configurations that are far from noisy obstacles will still be forced to satisfy difficult risk bounds. 
A few works have considered non-uniform risk allocation, either by formulating multi-stage optimization problems~\cite{ono08}, iteratively penalizing and relaxing risky waypoints from previous solutions~\cite{dai19}, or using differentiable surrogate risks encoded as variables in a nonlinear optimization problem~\cite{dawson2020}.
Similar to~\cite{blackmore06_aprobabilistic,lew20_chanceconstrained}, our method enforces joint chance constraints by using the union bound (Boole's inequality) and solving a set of individual chance constraints.
However, in our method, the risk bound of individual chance constraints for all obstacle-link pairs are decision variables in our optimization formulation.

Other methods design certificates that verify that a path is safe under a noise model.
\citet{vandeberg2010_lqgmp} assess path safety by assuming a linear-quadratic controller with Gaussian uncertainty (LQG-MP).
Several candidate paths are generated using a sampling-based planner and the best is chosen for execution.
\citet{axelrod18} certify a path as safe for a given level of risk if the robot's swept volume does not intersect a set of unsafe regions.
\citet{park2020_efficient} design probabilistic collision checkers for non-Gaussian distributions, which they use in an optimization-based planner to encourage safety. %
\citet{quinteropena2021_robustoptimization} also handle non-Gaussian distributions by solving a robustly formulated sequential convex programming problem.
\citet{dai19} generate candidate paths, propagate the uncertainty along the path using LQG-MP and estimate the resulting risk of collision via numerical integration.
Our proposed approach also generates risk-agnostic candidate paths which it then transforms into safe paths by solving a sequence of convex optimization problems.

\subsection{Implicit Representations and Uncertainty Quantification}
\label{sec:implicit_neural_representations_and_uncertainty_quantification}
Recent machine learning advances have produced efficient implicit neural representations of spatial information, such as Neural Radiance Fields (NeRFs)~\cite{mildenhall2020_nerf} and Signed Distance Fields~\cite{park2019_deepsdf}.
Robotics researchers have used these representations to learn multi-object dynamics~\cite{driess2022_learningmultiobject}, as manipulation planning constraints~\cite{driess2021_learning_models}, to achieve reactive robot manipulation~\cite{koptev2023_neural_joint} and to perform visual-only robot navigation~\cite{adamkiewicz2022_vision-only}.
Beyond their compact, efficient storage~\cite{adamkiewicz2022_vision-only}, these representations are advantageous for planning due to their continuous representation of geometry~\cite{adamkiewicz2022_vision-only,koptev2023_neural_joint,kurenkov2022_nfomp_neural} and ability to be learned directly from sensor data~\cite{camps2022_learning}.
Recent work has also investigated quantifying the uncertainty of a learned model~\cite{clements2019_risk,lahlou2023_deup,acharya2023_learningtoforecast}.
Methods to estimate both aleatoric and epistemic uncertainty have been proposed in the computer vision~\cite{kendall2017_what,vasconcelos2023_uncertainr,shen2021_stochastic_neural} and reinforcement learning~\cite{acharya2023_learningtoforecast,clements2019_risk,lahlou2023_deup} literatures.
This is important to enable the design of uncertainty-aware algorithms for downstream tasks.
For example,~\citet{shen2021_stochastic_neural,shen2022_conditional} propose probabilistic frameworks that attempt to capture uncertainty in a NeRF for synthetic novel view and depth-map estimation.
Similarly, our method takes a probabilistic approach to quantifying aleatoric sensing uncertainty.
However, our proposed neural representation also fuses kinematic information about a robot with spatial information to produce a robot-configuration-conditioned probabilistic distance model.\looseness=-1

\begin{figure}
    \centering
    \includegraphics[width=\linewidth]{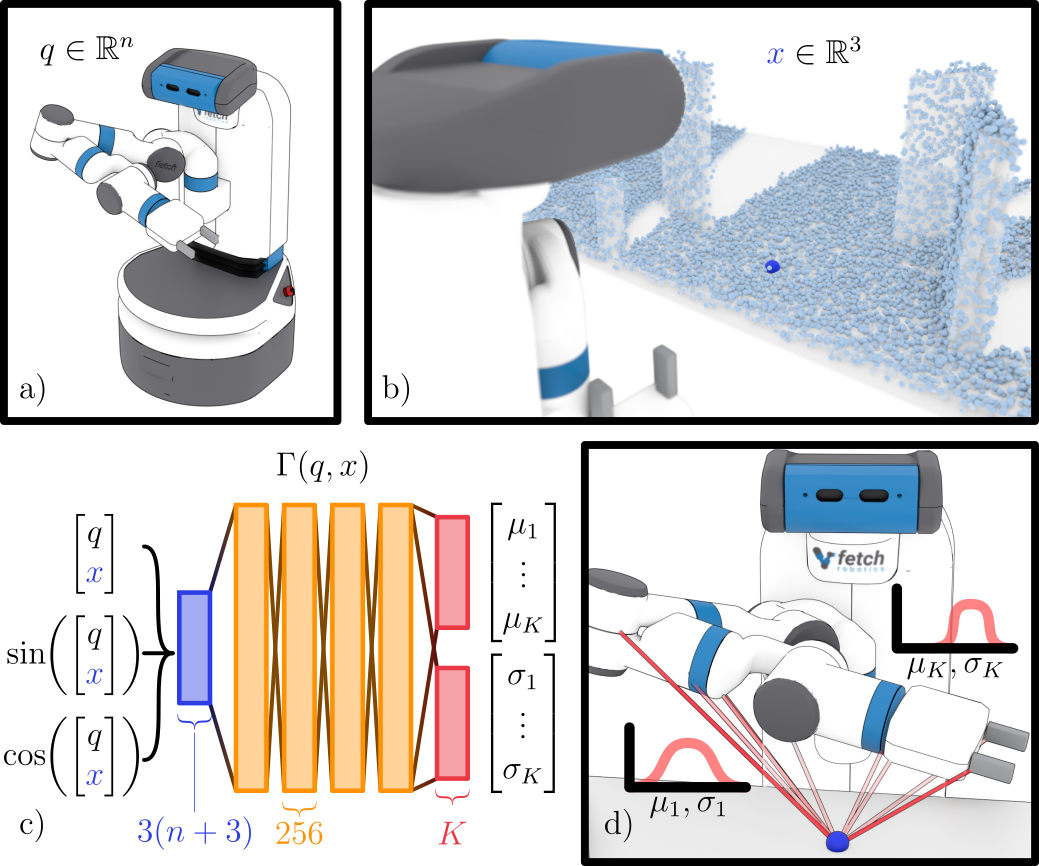}
    \caption{Our stochastic implicit neural signed distance functions uses \textbf{a)} a robot configuration $q$ and \textbf{b)} one noisy point $x$ as input. \textbf{c)} Inputs go through a positional encoding layer and then through $4$ fully connected layers of size $256$. Finally, two separate layers of size $K$ output the mean and standard deviation parameters of \textbf{d)} each link's distribution modeling the noisy signed distance conditioned on $q,x$.}\label{fig:planning_framework}
\end{figure}
\section{Safe Motion Planning with a Stochastic Neural Representation}\label{sec:technical-approach}

In this work, we assume that information about the environment is captured through a sensor as noisy $3$D points, akin to a point cloud. 
This noise is aleatoric from the perspective of the planner as it stems from immutable properties of the sensor and is irreducible.
We propose to quantify this aleatoric sensing uncertainty through a stochastic implicit neural representation that models noisy signed distances between the environment and the robot geometry.
Our neural representation, inspired by~\cite{koptev2023_neural_joint}, captures not only geometric information about the environment (as in work based on NeRFs~\cite{driess2022_learningmultiobject,adamkiewicz2022_vision-only} or SDFs~\cite{driess2021_learning_models,kurenkov2022_nfomp_neural,camps2022_learning}), but also kinematic information about the robot itself, which makes it suitable for motion planning for manipulation.
We find safe paths despite sensing errors by using this representation in a novel \emph{hierarchical motion planner}, instead of directly attempting to reformulate and solve~\ref{prob:prob1}.
Our planner first finds a candidate path using only the noisy sensed points (without knowledge of their noise), and then uses this candidate path and a user-provided bound on the risk of collision to compute a safe path.
The following sections describe our aleatoric sensing representation and planning framework.

\subsection{Stochastic Neural Implicit Signed Distance Representation}\label{sec:stochastic_neural_implicit_signed_distance_representation}
\citet{koptev2023_neural_joint} propose an implicit neural representation that models the signed distance between each robot link and arbitrary points in space.
The neural representation learns $\Gamma: \cspace \times \mathbb{R}^3 \rightarrow \mathbb{R}^K$ comprising $K$ related mappings $\Gamma_k: \cspace \times \mathbb{R}^3 \rightarrow \mathbb{R}$. 
Each $\Gamma_k(q,x)$ is the minimum distance function for the $k$-th robot link ($1 \le k \le K$), evaluated at the 3D point $x$ when the robot is in configuration $q$.
This representation is useful for motion planning for manipulation due to
\begin{enumerate*}[label=\arabic*)]
    \item representing distances to arbitrary points in the workspace without depending on specific geometry and
    \item its gradients point away from obstacles in \emph{configuration space}
\end{enumerate*}.

Inspired by this representation, we propose to learn a distribution, $S$, over signed distance functions, such that the distance between each robot link and points in the workspace is modeled as a Gaussian random variable.
We want to learn the posterior of $S$ conditioned on a training set $\mathcal{T}$ consisting of a finite collection of robot configurations $q_i$, 3D points $x_i$ and per-link noisy signed distance values $d_i^k$, \ie  $\mathcal{T} = \{\left ( \{d_i^k\}_{k=1}^{K}, q_i, x_i \right)\}_{i=1}^{N}$.

We formulate the problem using a Bayesian approach~\cite{shen2021_stochastic_neural} to compute the posterior $\prob{S | \mathcal{T}}$.
Note that explicitly computing this posterior is intractable since it would require the computation of the evidence \ie the marginal density of the observations.
Instead, we approximate it using variational inference (VI), where a parametric distribution~$\psi_{\theta}(S)$ approximates the true distribution.
The goal of VI is to find the parametric distribution that is closest to the true distribution, measured via their KL divergence~\cite{blei2017_variational}.
As the KL divergence is not computable because it requires the evidence, VI typically optimizes the evidence lower bound (ELBO)~\cite{blei2017_variational}.
For our problem, the VI formulation is:
\begin{subequations}
\label{eq:deterministic_nnik}
\begin{align}
& \underset{\theta}{\text{min}}
 \underbrace{\mathbb{E}_{\psi_{\theta}(S)}\log\left( \frac{\psi_{\theta}(S)}{p\left(S\right)} \right)}_{\text{KL-divergence prior}} - \underbrace{\mathbb{E}_{\psi_{\theta}(S)}\log\left( \prob{\mathcal{T}~|~S}\right)}_{\text{Log likelihood}}\; \label{eq:neg_elbow}
\end{align} 
\end{subequations}
where the first term is the KL divergence between $\psi_{\theta}$ and a prior $p(S)$ on the signed distance field, to encourage densities close to the prior, and the second term is the negative training set likelihood over the approximate posterior~$\psi_{\theta}$, which will choose parameters~$\theta$ that best explain the observed data.

We assume that $\psi_{\theta}$ can be factored as the product of independent Gaussian densities, $\psi^k_{\theta}(d|q,x)$, representing the distance fields for each robot link $k$.
These densities are jointly modeled as a neural network, $\Gamma(q,x)$, that outputs the parameters of $\psi_{\theta}$, $\{\mu_{1}, \sigma_{1},\dots,\mu_{K}, \sigma_{K}\}$ (see~\mf{fig:planning_framework} for network architecture).
The second term in~\mf{eq:neg_elbow} is computed in closed form using the likelihood of the Gaussian distribution.
For the first term we assume that (similar to $\psi_\theta$) the prior can be factored as a product of Gaussians, $p^k(d)$ with parameters $\{\mu^p_{k}, \sigma^p_{k},\}$.
The KL divergence between these distributions can be computed analytically as:
\begin{align*}
\text{KL}\left( \psi_{\theta}(S) || p \left(S \right) \right) &= 
   \sum_{k=1}^{K}\sum_{x \in \mathbb{R}^3}\sum_{q \in \mathbb{Q}}\text{KL}\left(\psi_{\theta}^k(S | x,q) || p^k \left(S \right) \right) \\
  &\approx \sum_{k=1}^{K}\sum_{i}\frac{\sigma_{k,i}^2 + (\mu^p_{k,i} - \mu_{k,i})^2}{2{\sigma^p_{k,i}}^2}   -\log \sigma_{k,i}
\end{align*}
In practice, we use a fixed number of samples to remove the dependency of the approximate posterior on $x$ and $q$.

\subsection{Chance-Constrained Hierarchical Planning}\label{sec:chance-constrained_layered_planning}
We propose a hierarchical motion planner to generate safe robot motions, described in~\mfa{algo:safe_motion_planning}.
First, an off-the-shelf motion planner~\cite{kavraki96_probabilistic,lavalle00_rapidly,kuffner00_rrtconnect,ratliff09_chomp,schulman2014_motion} is used to find a candidate path $\rho^c$ in the noisy sensed environment $\Xi$ (\mfa{line:motion_planning}) .
For each waypoint of $\rho^c$ (\mfa{line:waypoint_it}), we solve a chance-constrained \ik problem (\ccikop,~\mfa{line:ccikopt}) to compute the motion to the \emph{next} waypoint.
We use the pose of the robot's end-effector at $q^c_{j+1}$ as a soft constraint for the $j$-th \ik problem, encouraging solutions close to the original path.
We accumulate the risk allocated to each waypoint to ensure that it does not exceed the bound $\Delta$ for the total path (\mfa{line:remaining_risk}).
Each \ik problem is allowed up to the full remaining risk available, and returns an upper bound on the risk allocated to the corresponding waypoint.
This method can be seen as using $\rho^c$ as \emph{guidance} for the sequence of \ik problems, while flexibly accommodates the allowable risk bounds to compute a safe path, $\rho^s$.
\begin{algorithm}[t]
	\SetKwData{CandPath}{$\rho^c$}\SetKwData{RemainingRisk}{$\Delta_j^r$}\SetKwData{CCIKOut}{$(\Delta q, \delta, \gamma)$}\SetKwData{CCIKIn}{$(q^c_{j+1}, q^s_{j}, \Delta_j)$}
	\SetKwFunction{MotionPlanning}{MotionPlan}\SetKwFunction{CCIKOPT}{CCIKOPT}
	\SetKwInOut{Input}{input}\SetKwInOut{Output}{output}

	\Input{$q_{start}, \cspace_{goal}, \Xi, \Delta$}
	\Output{$\rho^s$, $\Delta_{T-1}$}
	\BlankLine

        \CandPath$\leftarrow$ \MotionPlanning$(q_{start}, \cspace_{goal}, \Xi)$\; \label{line:motion_planning}
        $\Delta_0 \leftarrow \Delta$, $q^s_0 \leftarrow q^c_0$\;
        \For{$j\leftarrow 0,\dots,T-1$}{\label{line:waypoint_it}
            \If{\CCIKOut$\leftarrow$\CCIKOPT\CCIKIn}{\label{line:ccikopt}
                $q^s_{(j+1)} \leftarrow q^s_j + \Delta q$\;\label{line:nextq}
                $\Delta_{j+1} \leftarrow \Delta_j - \gamma$\;\label{line:remaining_risk}
            }
            \Else{
                break\;
            }
        }
        \If{$j < T-1$}{\label{checkconvergence}
            return false\;
        }
        return $[q^s_0, \dots,q^s_T], \Delta_{T-1}$\;
	\caption{Chance-Constrained Hierarchical Motion Planner}\label{algo:safe_motion_planning}
\end{algorithm}\DecMargin{3em}

We extend the \ik formulation of~\cite{koptev2023_neural_joint} to the chance-constrained \ik setting in~\ref{prob:prob2}:
\begin{subequations}
\begin{align}
\tag{Prob. 2}\label{prob:prob2}
& \underset{\bm{\Delta q}, \bm{\delta}}{\text{min}} \nonumber
& & \bm{\Delta q^{T}} Q \bm{\Delta q} + \bm{\delta^{\top}}D\bm{\delta}\nonumber\\
& \text{s.t.}
& & q^{l} \leq q^s_j + \bm{\Delta q} \leq q^{u} \nonumber \\
& & & \text{FK}(q^s_j) + J^{\top}(q^s_j)\bm{\Delta q} = \text{FK}(q^c_{j+1}) + \bm{\delta} \nonumber \\
& & & \prob{\bigwedge_{r,k}  -\nabla \Gamma_{k,r}^{\top} \bm{\Delta q} \leq \Gamma_{k,r} - r_r} \geq 1 - \Delta_j \label{eq:original_joint_chance_constraint}
\end{align} 
\end{subequations}
with decision variables $\bm{\Delta q}$ and $\bm{\delta}$.
$\bm{\Delta q}$ corresponds to the robot motion between $q^s_j$ and $q^s_{j+1}$; $\bm{\delta}$ is a vector of slack variables that provide flexibility on the goal pose of the end-effector.
We minimize a quadratic function of the decision variables to encourage small motions that end close to the original end-effector pose from $\rho^c$.
Constraint~\eqref{eq:original_joint_chance_constraint} is the joint chance constraint requiring the risk of collision to remain under a given threshold $\Delta_j$ for all robot links $1 \le k \le K$ and noisy points $1 \le r \le R$.
In its deterministic version~\cite{koptev2023_neural_joint}, when $\Gamma_{k,r} = \Gamma_k(q^s_{j},x_r)$ becomes small, $\bm{\Delta q}$ is forced to align with $-\nabla \Gamma_{k,r}$ (which points away from collision with $r$) to avoid potential collisions.
In our approach, $\Gamma_{k,r}$ are random Gaussian variables, and we enforce that the probability that this constraint is satisfied is above a given threshold.
In the next section we describe our reformulation of the constraint to make the problem tractable.

\subsection{Reformulation of the Chance-Constrained IK Problem}
\label{sec:reformulation_of_chance_constrained_IK_problem}
For simplicity, let $\bm{x^{\top}} = [\bm{\Delta q^{\top}}, \bm{\delta^{\top}}]$, $A = \text{diag}([Q,~D])$, $B = [J^{\top}(q^s_j),~-I]$, $c^{\top} = [-\nabla \Gamma_{k,r}^{\top},~0 ]$, $g = \Gamma_{k,r} - r_r$, and $b = \text{FK}(q^c_{j+1})-\text{FK}(q^s_j)$.
We also use the following facts:\\
\begin{enumerate*}[label=\textbf{(4\alph*)},itemjoin={{, }}]
    \item $\prob{\bigwedge_i A_i} \geq 1 - p \iff \prob{\bigvee_i  \bar{A}_i} \leq p$ 
    \item $\prob{\bigvee_i A_i} \leq p \Longleftarrow \sum_i \prob{A_i} \leq p$
    \item $\sum_i \prob{A_i} \leq p \Longleftarrow \prob{A_i} \leq p_i, \; \forall i, \sum_i p_i \leq p$ 
\end{enumerate*}. 

\setcounter{equation}{4}

We rewrite~\eqref{eq:original_joint_chance_constraint} in simplified notation, then apply \textbf{(4a-c)}:
\begin{align*}
&\prob{\bigwedge_r\bigwedge_k c^{\top}\bm{x} \leq g} \geq 1-\Delta_j, &\\
&\Longleftarrow \sum_r \prob{ \bigvee_k  c^{\top}\bm{x} \geq g } \leq \Delta_j, & \text{\textbf{(4a)}, \textbf{(4b)}}\\
&\Longleftarrow \sum_k\prob{ c^{\top}\bm{x} \geq g } \leq \bm{y_r} \; \forall r, \; \sum_r \bm{y_r} \leq \Delta_j, & \text{\textbf{(4c)}, \textbf{(4b)}}\\
&\Longleftarrow \prob{c^{\top}\bm{x} \geq g} \leq \bm{\gamma_{k,r}} \; \forall r, k, \; \sum_{k,r}\bm{\gamma_{k,r}} \leq \Delta_j, & \text{\textbf{(4c)}}\\
&= \prob{c^{\top}\bm{x} \leq g } \geq 1 - \bm{\gamma_{k,r}} \; \forall r, k, \; \sum_{r,k}\bm{\gamma_{k,r}} \leq \Delta_j&
\end{align*}
where $\bm{\gamma_{k,r}}$ is the risk allocated to link $k$ and point $r$.
By properties of the Gaussian CDF~\cite{prekopa1995_stochastic}, we know $\prob{a^{\top}b \leq c} \geq p \iff a^Tb-\mu_c+\sigma_c\phi^{-1}(p) \leq 0$ for $c \sim \mathcal{N}(\mu_c, \sigma^2_c)$, where $\phi^{-1}$ is the inverse CDF of the standard normal distribution.
Thus, we can write the following deterministic reformulation for~\eqref{eq:original_joint_chance_constraint}:

\begin{align}
    c^{\top}\bm{x}-\mu_{k,r}+\sigma_{k,r}\phi^{-1}(\bm{\bar{\gamma}_{k,r}}) &\leq 0 \; \forall k,r \label{eq:reformulated_constraint1}\\
    \sum_{k,r}(1-\bm{\bar{\gamma}_{k,r}}) &\leq \Delta_j\label{eq:reformulated_constraint2}
\end{align}
where $\bm{\bar{\gamma}_{k,r}} = 1-\bm{\gamma_{k,r}}$.
Finally, we provide a conservative reformulation of~\eqref{eq:reformulated_constraint1} by noting that, for $0.5 \leq x < 1$, $\sqrt{\pi/8}\log\left( x/(1-x) \right) \geq \phi^{-1}(x)$, allowing:

\begin{equation}
 \text{\eqref{eq:reformulated_constraint1}}\Longleftarrow c^{\top}\bm{x}-\mu_{k,r}+\sigma_{k,r}\sqrt{\frac{\pi}{8}}\log\left( \frac{\bm{\bar{\gamma}_{k,r}}}{1-\bm{\bar{\gamma}_{k,r}}} \right) \leq 0 \label{eq:deterministic_reformulated_chance_constraint}
\end{equation}
which requires the risk variables to be in $0 < \bm{\gamma_{k,r}} \leq 0.5$.
This restriction is reasonable in our context since we are interested in paths with low collision risk.
Using~\eqref{eq:reformulated_constraint2} and~\eqref{eq:deterministic_reformulated_chance_constraint} instead of~\eqref{eq:original_joint_chance_constraint} allows us to write the following optimization problem:\looseness=-1
\begin{align*}
\tag{Prob. 3}\label{prob:prob3}
& \underset{\bm{x}, \bm{\bar{\gamma}}}{\text{min}}
& & \bm{x}^{\top} A \bm{x} + h^{\top}\bm{\bar{\gamma}}\\
& \text{s.t.}
& & x_l \leq \bm{x} \leq x^u \;\\
& & & \bar{\gamma}_l \leq \bm{\bar{\gamma}} \leq \bar{\gamma}^u \;\\
& & & B\bm{x} = b \;\\
& & & c^{\top}\bm{x}-\mu_{k,r}+\sigma_{k,r}\sqrt{\frac{\pi}{8}}\log\left( \frac{\bm{\bar{\gamma}_{r,k}}}{1-\bm{\bar{\gamma}_{r,k}}} \right) \leq 0,\; \forall k,r \\
& & &\sum_{r,k}(1-\bm{\bar{\gamma_{k,r}}}) \leq \Delta_j
\end{align*} 

We have added a linear term on the objective function of the reformulated problem to minimize the amount of risk allocated to the waypoint at the $j$-th iteration.
This per-waypoint minimum risk behavior of our formulation is necessary to account for potentially high-risk future waypoints on the path.
It also has the effect of allowing us to \emph{globally} minimize (with respect to $\Gamma$ and $\rho^c$) the risk of $\rho^s$.
To solve~\ref{prob:prob3}, we create piecewise-affine conservative approximations of the $\log$ functions in the collision risk chance-constraints.
This approximation creates mixed-integer programs that can be solved using commercial solvers~\cite{gurobi} to global optimality at the cost of potentially high computation time.
\section{Evaluation and Results}
\label{sec:experiments}
We evaluate our proposed approach on a $n=8$~\dof Fetch robot with $K=11$ links, corresponding to those in the kinematic chain of its end effector (including the torso and fingers).
We use PyBullet~\cite{coumans2021_pybullet} for collision checking, PyTorch~\cite{paszke2019_pytorch} for neural network training, OMPL's Python bindings~\cite{sucan2012_ompl} for planning and Gurobi~\cite{gurobi} as our optimizer.
All experiments were conducted on an Intel i7-12700K CPU and a RTX2080Ti GPU.

\subsection{Implicit Neural Representation}\label{sec:results_stochastic_signed_distance_neural_representation}
We parameterize our implicit stochastic distance model as a feed-forward neural network (\cref{fig:planning_framework}c).
For a $n$-\dof robot, the network takes an input tensor of size $3 * (n + 3)$ comprising the $n$ values of the robot configuration concatenated with the $3$ coordinates of the environment point, as well as the sine and cosine of these values. 
These trigonometric components serve as a form of positional encoding similar to that used in standard neural radiance fields~\cite{mildenhall2020_nerf}.
The network has a shared core of four fully connected 256-wide layers with rectified linear unit (ReLU) activation. 
For a robot with $K$ links, the output of these layers is used (independently) with one additional fully connected layer of size $256 \times K$ to predict the mean distance from the environment point to each link's geometry, as well as with another fully connected layer of size $256 \times K$ and a softplus layer of size $K$ to predict the standard deviation of these distances.

We generate a dataset of noisy distance samples from a simulated sensor to train the distance model. 
Similarly to~\citet{koptev2023_neural_joint}, we sample a set of robot configurations ($Q$) uniformly at random. 
For each configuration, we sample a set of environment points uniformly at random ($P_R$), a set of environment points \emph{near} to each link ($P_N$) and a set of environment points \emph{inside} each link ($P_I$). 
We compute the true shortest distance between each point and link using PyBullet.
We then simulate a set of noisy sensor measurements ($NS$) with mean at the true distance for each environment point and a fixed standard deviation ($\sigma$).
In our experiments, $|Q| = 3000$, $|P_R| = 500$, $|P_N| = K * 10$, $|P_I| = K * 20$, $|NS| = 50$, and $\sigma = \SI{2}{\cm}$.
This results in a total of 2.49 million sampled points, each of which has 50 noisy distance samples.
Empirically, this dataset is roughly balanced between points in collision and points in free space.\looseness=-1

We train the model on the collected dataset for $500$ epochs with an Adam~\cite{kingma2014adam} optimizer, learning rate of $1\times 10^{-4}$, and batch size of $512$.
We verify its performance by predicting distance distributions between robot links and a set of randomly generated 3D points from the waypoints of $1000$ discretized paths.
The gripper link shows an average error of $\SI{1}{\cm}$ for mean and $\SI{3.7}{\mm}$ for standard deviation while the elbow attains $\SI{0.7}{\mm}$ and $\SI{0.3}{\mm}$, respectively.
\mfa{fig:neural_network_results} shows the predicted and true distribution parameters for one path, one randomly selected point and these two robot links.

\begin{figure}
    \centering
    \includegraphics[width=0.99\linewidth]{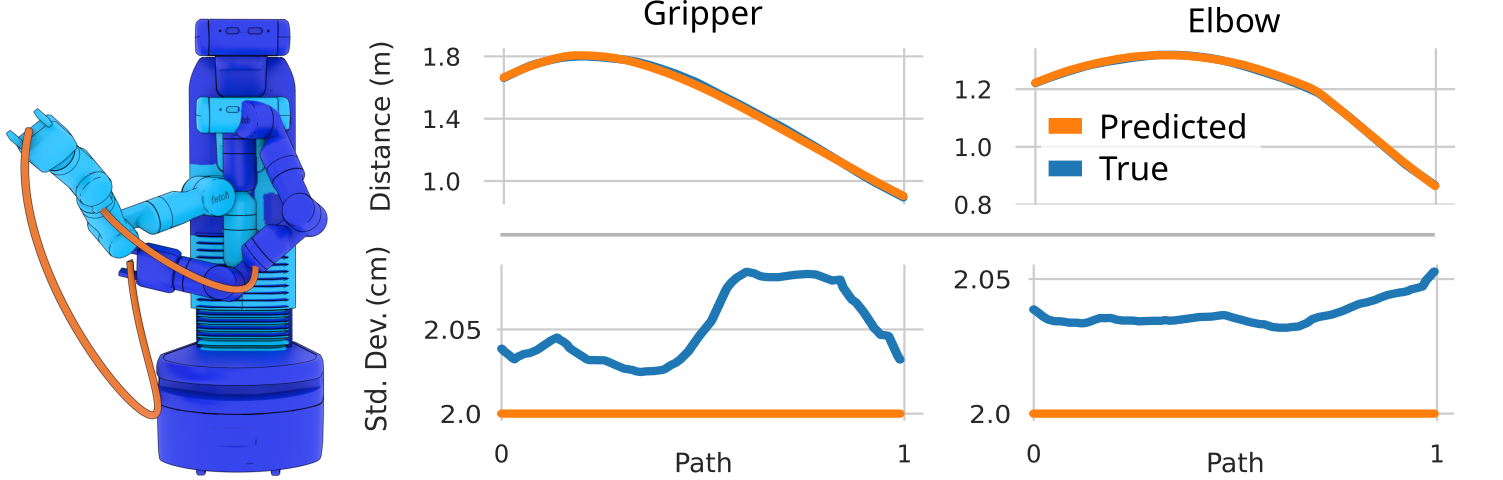}
    \caption{Comparison between true and predicted probability distribution parameters (mean at the top and standard deviation at the bottom) for the distance between robot links and a randomly generated point in space for the Gripper (left) and Elbow (right) links along a path of $500$ waypoints. Note that---despite the visual gap---the actual error in standard deviation is small, less than \SI{1}{\mm}.}
    \label{fig:neural_network_results}
\end{figure}

\subsection{Safe Motion Planning with Implicit Neural Representation}
We evaluate our proposed approach on a set of simulated tabletop manipulation problems generated using \mnet~\cite{chamzas2022-motion-bench-maker}.
The Fetch robot needs to plan to grasp an object, avoiding collisions with the table and obstacles upon it (\mfa{fig:planning_framework2}).
We create 50 problems by randomly perturbing the positions ($\pm 2.5$cm in $x,y,z$) and orientations ($\pm15^{\circ}$) of the objects of a nominal scene and the relative pose of the robot's base ($\pm10$cm in $x, y, z$ and $\pm90^{\circ}$) with respect to the table.
The environment is represented as a point cloud-like set of noisy 3D spheres of different radii that covers the (unknown to the planner) collision geometries of all objects.
We assume that the table's geometry is noise-free while the objects on top are noisily sensed, per~\mfa{sec:results_stochastic_signed_distance_neural_representation}.
Note that these problems were designed by~\citet{chamzas2022-motion-bench-maker} to be challenging and ``realistic'' from the motion planning perspective and require the robot to plan long, elaborate paths that need to avoid the table and then dodge collisions with the objects on top.

We compare the performance and safety of our approach with a commonly used baseline: inflating the environment's geometry to encourage the computation of paths that maintain larger clearance and have therefore less chances of colliding.
We inflate each sphere by increasing its radius by $20\%, 40\%,$ or $60\%$.
We also include results of $0\%$ inflation as a baseline to show the performance of a planner that is unaware of the sensing uncertainty.
Motion plans for all baselines, as well as the candidate paths used by our method, are computed using RRT-Connect~\cite{kuffner00_rrtconnect} with simplification enabled to encourage short and smooth paths.\looseness=-1

We estimate the risk of collision for each computed path using Monte-Carlo sampling with $20,000$ samples, where each sample draws sphere poses from the noisy sensed distribution.
For our method, we also show the guaranteed path-wise risk bound (Risk Bound) and estimated risk of collision of the candidate path before optimization (Initial Risk).
All problems have a maximum number of $15$ attempts to find any valid (\ie collision-free with respect to the inflated obstacles, for the baselines) plan.
The results are shown in~\mfa{fig:risk}.\looseness=-1

\begin{figure}
	\includeinkscape[width=\linewidth]{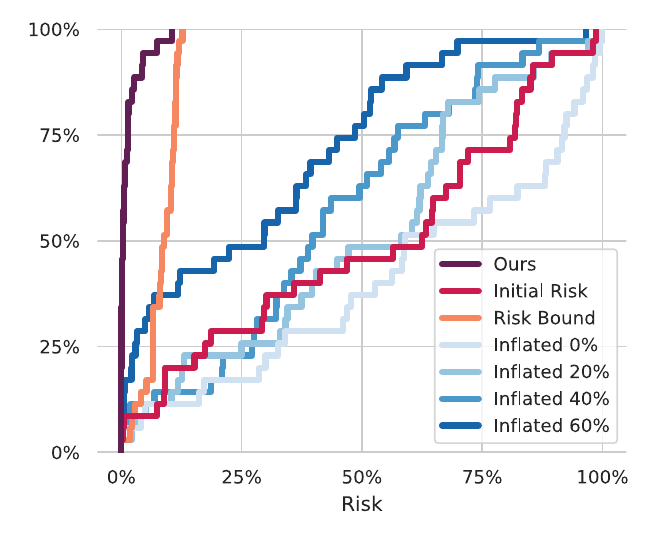}
 \vspace{-9mm}
 \caption{Estimated CDF of the risk attained by each method on all 50 problems.}
 \label{fig:risk}
 \vspace{-4mm}
 \end{figure}

We note that the uncertainty-unaware planner produces paths with the highly variable risk of collision (an average of $60\%$), which is likely unacceptable for safety-critical applications.
For higher parameter values of the inflated baseline, the estimated risk of collision decreases as expected due to a larger \cspacespace obstacle region that encourages larger clearance with the true geometry.
However, there is no clear relation between the inflation increase and the drop in risk which makes the baselines difficult to tune when a desired level of risk is required (see also~\mfa{tbl:results}).
Additionally, we note that success rate (not shown here) for the baseline methods started dropping significantly as the inflation ratio increased, suggesting a potential limit on minimum risk that they can attain for these problems.
We give our planner a maximum allowable risk bound of $10\%$ and ask it to return the minimum risk for each waypoint.
Our proposed approach can compute paths with significantly lower risk for most problems, starting from risky candidate paths.\looseness=-1

For each problem and method we also compute the path length and end-effector displacement as path quality metrics, as well as the time taken by the planner.
The results are summarized in~\mfa{tbl:results}.
\begin{table}[]
\tiny
    \centering
    \begin{tabularx}{\linewidth}{r || c c c c}
Method            & EE Disp. (m)         & Path Length (rad.)   &  Path Risk      & Planning Time (s)    \\
\hline
\hline
Inflated~0\%        & $2.32\pm0.82$        & $7.91\pm2.53$        & $0.60\pm0.33$        & $3.51\pm10.42$      \\
Inflated~20\%        & $2.39\pm0.76$        & $8.34\pm2.74$        & $0.48\pm0.29$        & $3.61\pm9.55$       \\
Inflated~40\%        & $2.33\pm0.79$        & $8.27\pm2.86$        & $0.42\pm0.26$        & $5.85\pm13.73$      \\
Inflated~60\%        & $2.38\pm0.85$        & $8.57\pm3.18$        & $0.28\pm0.25$        & $5.69\pm12.95$      \\
\hline
Proposed             & $2.16\pm0.60$        & $6.93\pm2.26$        & $0.01\pm0.02$        & $150.63\pm55.38$     \\
    \end{tabularx}
    \caption{Mean and standard deviation of path quality metrics for all methods on the 50 problems.}
    \vspace{-5mm}
    \label{tbl:results}
\end{table}
The table shows mean and standard deviation for each method over all $50$ problems.
The large values of path length and end-effector displacement are evidence of the high complexity of the computed paths due to the challenging motion planning problems.
Our method finds paths with lowest end-effector displacement and path length, which is the result of the minimization of motion displacement in our planner.
However, our method shows planning times that are orders of magnitude larger than the baselines.
This is mostly due to a large number of risk constraints being added to each chance-constrained problem, which creates large mixed-integer programs that require deep branch and bounds searches to find optimal solutions.
Despite this, it is noteworthy that our method can find paths with the lowest collision risk among the baselines without sacrificing path quality.\looseness=-1
\section{Concluding Remarks}
\label{sec:conclusion}
This paper presents a novel approach to planning under sensing uncertainty for high \dof robots that reliably computes safe paths without strong assumptions on the true environment geometry.
Our planner relies on an implicit neural representation trained to capture aleatoric uncertainty arising from the robot's sensor.
Our representation does not place assumptions on the environment but instead directly approximates signed distance distributions between the robot and points in space, conditioned on robot configurations.
We further show how this representation can be integrated with a hierarchical planner to compute paths with guaranteed bounds on the probability of collision (up to the quality of the model).
We have experimentally validated the merits of our approach on challenging, realistic manipulation motion planning problems to show that our method is capable of finding safe paths despite sensing uncertainty without reducing path quality.
As future work we will investigate how to further reduce the need for conservative over-approximations in our approach, since this will allow us to solve more tightly constrained motion planning problems, such as those found in manipulation in clutter.
We will also seek to reduce the time taken by our method, in part by applying intelligent constraint subset selection heuristics~\cite{hauser_semiinfinite_2021} to simplify the optimization problems solved at each waypoint.
Finally, we will further investigate the need to consider the epistemic uncertainty coming from our neural representation for planning; a problem that has recently gained much attention in machine learning~\cite{acharya2023_learningtoforecast}.

\newpage
\printbibliography{}

\end{document}